\pdfoutput=1
\documentclass[11pt,a4paper]{article}
\usepackage[final]{acl}

\usepackage{times,latexsym}
\usepackage{url}
\usepackage[T1]{fontenc}
\usepackage{CJKutf8}
\usepackage[utf8]{inputenc}
\usepackage{devanagari}
\usepackage{balance}

\title{MultiCoNER: A Large-scale Multilingual dataset for Complex Named Entity Recognition}

\author{Shervin Malmasi, Anjie Fang\thanks{~~These authors contributed equally in this work.}~~, Besnik Fetahu$^*$, Sudipta Kar$^*$, Oleg Rokhlenko \\
 Amazon.com, \\ Seattle, WA, USA \\
  
\texttt{{\{malmasi,njfn,besnikf,sudipkar,olegro\}@amazon.com}}}

\date{February 2022}
\usepackage{appendix}
\usepackage{pdfcomment}
\usepackage{cleveref}
\usepackage{graphicx}
\usepackage{xspace}
\usepackage{booktabs}
\usepackage{multirow}
\usepackage{soul}
\usepackage{enumitem}

\usepackage[OT2,T1]{fontenc}
\usepackage[utf8]{inputenc}
\usepackage{CJKutf8}

\newcommand{\ie}{i.e.,~}
\newcommand{\eg}{e.g.,~}

\newcommand{\example}[1]{\texttt{\small{#1}}}
\newcommand{\exampleq}[1]{``\texttt{\small{#1}}''}
\newcommand{\classname}[1]{\texttt{#1}}
\newcommand{\langid}[1]{\texttt{#1}}
\newcommand{\exampletagged}[2]{\textcolor{red}{[#1]}$_{\textcolor{blue}{{#2}}}$}

\newcommand\textcyr[1]{{\fontencoding{OT2}\fontfamily{wncyr}\selectfont #1}}

\newcommand{\mconer}{\textsc{MultiCoNER}\xspace}

\newcommand{\lowner}{\textsc{Lowner}\xspace}

\newcommand{\msquestion}{\textsc{Msq-Ner}\xspace}
\newcommand{\msquery}{\textsc{Orcas-Ner}\xspace}

\newcommand{\conll}{CoNLL03\xspace}
\newcommand{\ontonotes}{OntoNotes\xspace}

\newcommand{\xlmr}{B}
\newcommand{\gemnet}{GM}

\newcommand{\zh}[1]{\texttt{\begin{CJK*}{UTF8}{gbsn}#1\end{CJK*}}}

\newcommand{\change}[1]{\textcolor{black}{#1}}

\begin{document}

\maketitle

% train tokens: 3134950
% dev tokens: 144641
% test tokens: 23051259. 
% total: 26,330,850 

\begin{abstract}
We present \mconer, a large multilingual dataset for Named Entity Recognition that covers 3 domains (Wiki sentences, questions, and search queries) across 11 languages, as well as multilingual and code-mixing subsets.
This dataset is designed to represent contemporary challenges in NER, including low-context scenarios (short and uncased text), syntactically complex entities like movie titles, and long-tail entity distributions.
The 26M token dataset is compiled from public resources using techniques such as heuristic-based sentence sampling, template extraction and slotting, and machine translation.
%
% We evaluated the data with two NER models: a baseline XLM-RoBERTa model, and a state-of-the-art NER GEMNET model that leverages gazetteers.
We applied two NER models on our dataset: a baseline XLM-RoBERTa model, and a state-of-the-art GEMNET model that leverages gazetteers.
%, measuring their overall macro-F1 and per-class NER performance.
The baseline achieves moderate performance (macro-F1=54\%), highlighting the difficulty of our data.
%Addition of gazetteers in  GEMNET yields a significant
GEMNET, which uses gazetteers, improvement significantly  (average improvement of macro-F1=+30\%).
%s have a poor cross-domain and cross-lingual performance.
% The addition of gazetteers in GEMNET yields a significant improvement across all tracks, with an average improvement of macro-F1=+30\%.
%
\mconer poses challenges even for large pre-trained language models, and we believe that it can help further research in building robust NER systems.

\mconer is publicly available,\footnote{\tiny\url{https://registry.opendata.aws/multiconer/}} and we hope that this resource will help advance research in various aspects of NER.

\end{abstract}

\section{Introduction}

Named Entity Recognition (NER) is a core task in Natural Language Processing which involves identifying entities in text, and recognizing their types (\eg classifying entities as a person or location).
Recently, the development of Transformer-based NER approach have results in new state-of-the-art (SOTA) results on well-known benchmark datasets like \conll and \ontonotes%~\citep{huang2015bidirectional,devlin19bert}.
\citep{devlin19bert}.
Despite these strong results, there remain a number of practical challenges that may not be represented by these existing datasets.

As noted by \citet{augenstein2017generalisation}, increasingly higher scores on these datasets are driven by several factors:

\begin{figure}
    \centering
    \includegraphics[width=\columnwidth]{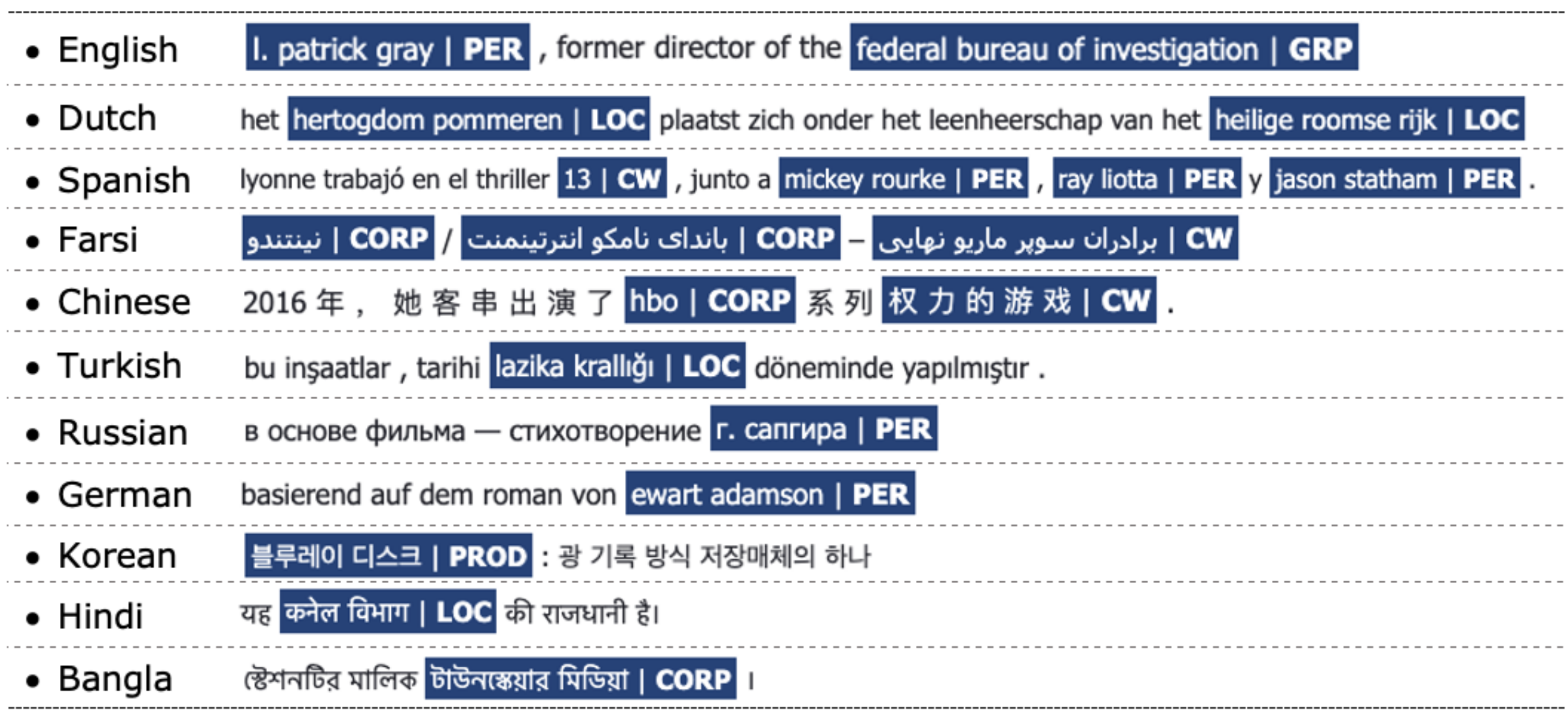}
    \caption{Some examples for all the languages incorporated in \mconer.}
    \label{fig:front-page-img}
\end{figure}

\begin{itemize}[leftmargin=*]
\itemsep0em 
    \item Well-formed data, with punctuation and capitalized nouns, makes the NER task easier, providing the model with additional contextual cues.
    
    \item Texts from articles and the news domain contain long sentences with rich context around entities, providing valuable signals about the boundaries and types of entities.
    
    \item Data from the news domain\footnote{e.g. \conll~\cite{sang2003introduction}} contains ``easy'' entities such as country, city, and person names, allowing pre-trained models to perform well due to their existing knowledge of such tokens.
    
    \item Memorization effects, due to large overlap of entities between the train and test sets also increases performance.
    
\end{itemize}

Accordingly, models trained on existing benchmark datasets such as \conll tend to perform significantly worse on unseen entities or noisy text~\cite{DBLP:conf/naacl/MengFRM21}.

\pagebreak

\subsection{Contemporary Challenges in NER}
\label{sec:ner_challenges}

There are many challenging scenarios for NER outside of the news domain. We categorize the challenges typically encountered in NER according to several dimensions: (i) available context around entities, (ii) named entity surface form complexity, (iii) frequency distribution of named entity types, (iv) dealing with multilingual and code-mixed textual snippets, and (v) out-of-domain adaptability.

\paragraph{Context}
News domain text often features long sentences that reference multiple entities. In other applications, such as voice input to digital assistants or search queries issued by users, the input length is constrained and the context is less informative. 
Datasets featuring such low context settings are needed to assess model performance.

Additionally, capitalization and punctuation features are large drivers of success in NER \citep{mayhew2019ner}. However, inputs such as search queries from users, or voice commands transcribed using ASR, lack these surface features. To understand model performance in such cases, an uncased dataset is needed.

\paragraph{Entity Complexity}
Datasets in existing NER benchmarks are often dominated by entities representing persons, locations, and organizations. Such entities are often composed of proper nouns, or have names with simple syntactic structures.
However, not all entities are so simple in structure: some entity types (\eg creative works) can be linguistically complex.
They can be
complex noun phrases (\example{Eternal Sunshine of the Spotless Mind}),
gerunds (\example{Saving Private Ryan}), 
infinitives (\texttt{\small{To Kill a Mockingbird}}), 
or full clauses (\example{Mr. Smith Goes to Washington}).
Syntactic parsing of such nouns is hard, and most current parsers and NER systems fail to recognize them. 
The top system from WNUT 2017 achieved $8\%$ recall for creative work entities \citep{aguilar2017multi}.
Corpora including such challenging entities are needed for evaluation of model performance in such cases.

\paragraph{Entity Distributions}

In many domains, entities have a large long-tail distribution, with millions of possible values (\eg location names). This makes it hard to build representative training data, as a small dataset can only cover a portion of the potentially infinite entity space. A very large test set is required for comprehensive evaluation.

Furthermore, some domains have entity spaces that are continuously growing.
While all entity types are open classes (\ie new ones are added), some groups have a faster growth rate, \eg new books, songs, and movies are released weekly. Assessing true model generalization requires test sets with many entities that are unseen in the training set, in order to mimic an open-world setting.

\paragraph{Multilinguality and Code-Mixing}
The recent success of multilingual models have greatly boosted task performance in languages with fewer resources, by leveraging transfer learning from high resource languages.
However, there are limits to what can be achieved with cross-lingual transfer, and training data in additional languages is necessary to make progress in this field.
The availability of a NER dataset that addresses all the above challenges across many languages will enable new research directions in multilingual model evaluation, as well as for few- and zero-shot cross-lingual transfer scenarios.

Code-mixing, where entities and the main text may be in different languages, is another related research area in multilingual NER where additional resources can help. Code-mixed data is increasing online, especially in social media platform where multiple languages are used in a single post. Such a dataset is needed to study this area and evaluate truly multilingual NER systems. 

\paragraph{Domain Adaptation} A robust NER model is expected to perform effectively in several domains, such as well written sentences, questions, web search queries, etc. While well written sentences can be easy for NER, shorter questions and queries can be challenging. It is important to study how to adapt existing NER into newly emerging domains. However, most of the existing NER benchmarks only focus on data in a single domain limiting its usage for studying domain adaptation. 

%\subsection{Contributions}

\paragraph{\mconer} We address the aforementioned challenges by presenting \mconer, a 
% \textbf{Multi}lingual \textbf{Co}mplex \textbf{N}amed \textbf{E}ntity \textbf{R}ecognition dataset 
multilingual dataset that features a large number of entities (including complex ones) in three distinct domains that represent different challenges. Some key facts about \mconer:

\begin{itemize}[leftmargin=*]
    \item Textual snippets in \mconer are low in context, allowing to assess NER model's capability in detecting ambiguous named entities;
    \item Named entities contain a highly diverse distribution from simple \classname{Location} (\classname{LOC}) to highly complex entities \classname{Creative Work} (\classname{CW});
    \item Using open knowledge bases such as Wikipedia and Wikidata, we generate textual snippets that contain highly diverse named entities;
    \item Through a combination of localized versions of Wikipedia and Wikidata, and as well as automated text translation approaches, we generate NER data for 11 languages and 3 domains that can be used to test cross-lingual and cross domain NER performance. Some examples are presented in Figure \ref{fig:front-page-img}.
\end{itemize}

%To evaluate the effectiveness of our model, we ...

%\section{Related Work}
%\TK ...

\section{\mconer Dataset Overview}

The \mconer dataset was designed in order to address the NER challenges described in \S\ref{sec:ner_challenges}. It represents 3 domains (wiki sentences, questions, and search queries) and includes 11 languages, including multilingual and code-mixed subsets. \mconer was collected and released as part of the SemEval 2022 Task\#11, serving more than 236 participants across all the different languages~\cite{DBLP:conf/semeval/MalmasiFFKR22}.

\begin{table}
\centering
\resizebox{1\columnwidth}{!}{
\begin{tabular}{p{3.8cm}|p{7.7cm}}
\toprule
Source & Gold Sentence \\
\midrule
English -- Wiki  & \exampletagged{heat vision and jack}{CW}, a 1999 television pilot \\
Spanish -- Wiki & reina consorte de \exampletagged{escocia}{LOC} como esposa de \exampletagged{jacobo v}{PER}.\\
English -- QA &  when was the \exampletagged{nokia 2.2}{PROD} released \\
English -- Search Query &  cast of \exampletagged{dr. devil and mr. hare}{CW} \\
Russian -- QA & \textcyr{где было \exampletagged{королевы крика}{CW} снято} \\
%Korean -- Search Query & \begin{CJK}{UTF8}{}\CJKfamily{mj}에 대 한 일기 예보 \exampletagged{카보산루카스}{LOC}\end{CJK}\\
Code-Mixed (KO/EN) & \exampletagged{symphony no. 7 in e major}{CW} \begin{CJK}{UTF8}{}\CJKfamily{mj}란 무엇입니까?\end{CJK} \\
\bottomrule
\end{tabular}
}
\caption{\small{Examples from \mconer: entities are in brackets, followed by their type.}}
\label{tab:examples_v3}
\end{table}

\subsection{NER Taxonomy}
\label{sec:taxonomy}

\mconer leverages the WNUT 2017 \citep{wnut2017} taxonomy entity types, which defines the following NER tag-set with 6 classes:

\begin{itemize}[leftmargin=0.6em]
\itemsep0em 
    \item \classname{\small PERSON} (\classname{\small PER} for short, names of people)
    
    \item \classname{\small LOCATION} (\classname{\small LOC}, locations/physical facilities)
    
    \item \classname{\small CORPORATION} (\classname{\small CORP}, corporations/businesses)
    
    \item \classname{\small GROUPS} (\classname{\small GRP}, all other groups)
    
    \item \classname{\small PRODUCT} (\classname{\small PROD}, consumer products)
    
    \item \classname{\small CREATIVE-WORK} (\classname{\small CW}, movie/song/book titles)
\end{itemize}

This taxonomy allows us to capture a wide array of entities, including those with more complex entity structure, such as creative works.

\subsection{Languages and Subsets}

% \begin{itemize}
%     \item Bangla (BN)
%     \item Chinese (ZH)
%     \item Dutch (NL)
%     \item English (EN)
%     \item Farsi (FA)
%     \item German (DE)
%     \item Hindi (HI)
%     \item Korean (KO)
%     \item Russian (RU)
%     \item Spanish (ES)
%     \item Turkish (TR)
% \end{itemize}

\begin{table}[ht!]
\centering
\resizebox{.8\columnwidth}{!}{
\begin{tabular}{ l l | l l | l l }
\toprule
 Bangla &  (\langid{BN}) &  Hindi  & (\langid{HI})  & German  &  (\langid{DE}) \\ 
 Chinese &  (\langid{ZH}) & Korean  & (\langid{KO}) & Turkish & (\langid{TR}) \\  
 Dutch   &  (\langid{NL}) & Russian & (\langid{RU}) & Farsi   &  (\langid{FA})   \\
 English &  (\langid{EN}) & Spanish & (\langid{ES}) \\
\bottomrule
\end{tabular}}
\caption{\small{The languages included in \mconer, along with their 2-letter codes.}}
\label{tab:languages}
\end{table}

There are 11 languages included in \mconer (cf. \Cref{tab:languages}).
These languages were chosen to include a diverse typology of languages and writing systems, and range from well-resourced (\eg \langid{EN}) to low-resourced ones (\eg \langid{FA}).

\mconer contains 13 different subsets: 11 monolingual subsets for the above languages, a multilingual subset (denoted as \langid{MULTI}), and a code-mixed one (\langid{MIX}).

\paragraph{Monolingual Subsets}
Each of the 11 languages has their own subset with data from all domains.

\paragraph{Multilingual Subset}
This contains randomly sampled data from all the languages mixed into a single subset. This subset is designed for evaluating multilingual models, and should ideally be used under the assumption that the language for each sentence is unknown. A more detailed description of the multilingual train/dev/test set construction is provided in \S\ref{sec:data_construction}.

\paragraph{Code-mixing Subset}
This subset contains code-mixed instances, where the entity is from one language and the rest of the text is written in another language. Like the multilingual subset, this subset should also be used under the assumption that the languages present in an instance are unknown.

\subsection{Domains and Data Sources}

The three domains\footnote{Domain can have ambiguous interpretations~\cite{DBLP:conf/acl/WeesBWM15}, in our case it reflects a combination of provenance and text genre.} of \mconer are listed below. Details on the construction of the different subsets are provided in \S\ref{sec:data_construction}.

\paragraph{Wikipedia Sentences (\lowner)}
This subset of \mconer, which we call Low-context Wikipedia NER (\lowner) set, is built by sampling sentences from Wikipedia and using heuristics to identify ones that represent the NER challenges we target. More details on how we select sentence from Wikipedia are provided in \S\ref{subsec:wiki_sentences}.

\paragraph{Questions (\msquestion)}
The \msquestion subset of \mconer represents NER in the QA domain.
It is composed of a set of natural language questions, based on the MS-MARCO QnA corpus (V2.1)~\citep{bajaj2016ms}.

\paragraph{Search Queries (\msquery)}
The \msquery subset of \mconer represents the search query domain. %
To build this data, 
we utilize $10$ million Bing user queries from the ORCAS dataset \citep{craswell2020orcas}.

\subsection{Data Splits}
To ensure that obtained NER results on this dataset are \emph{reproducible}, we create three predefined sets for training, development and testing. The entity classes within each set are approximately uniformly distributed. Table~\ref{tab:data_stats} shows detailed statistics for each of the 13 subtasks and data splits.

% We create three sets for each subset:  training, development, and test. %These which are sampled from the extracted datasets. 
% The distribution of the entity classes within the sets is approximately balanced. Table~\ref{tab:data_stats} shows the dataset statistics for each of the 13 subsets, separated into their corresponding training, development, and test splits. 

\paragraph{Training Data} For the training data split, we limit the size to be $15,300$ sentences. The number of instances was chosen to be comparable to well-known NER datasets such as \conll~\cite{sang2003introduction}. Majority of the instances come from the \lowner domain, with a small sample of $100$ instances from the \msquestion and \msquery domains. These instances represent out-of-domain adaptation. The out-of-domain instances are limited in order to be able to realistically assess the out-of-domain performance of models trained on the \mconer dataset.

Note that in the case of the Multilingual subset, the training split contains all the instances from the individual language splits. For Code-Mixed on the other hand, we constructed only a small training split, in this way we allow for NER models to better model this task, rather than having abundance of code-mixed instances. The Code-Mixed instances are constructed by first sampling instances from the language specific training splits, and then at random replacing the original entity surface forms into their corresponding surface forms in another language present in our dataset.

\paragraph{Development Data} We randomly sample $800$ instances per subset from the \lowner domain (5\% of the training set size), a reasonable amount of data for assessing model generalizability.

The only difference in the development data is for the Multilingual and Code-Mixed subtastks, where the development splits are constructed similar  as for the training splits (see above).

\paragraph{Test Data} Finally, the testing set represents the remaining instances that are not part of the training or development set. To avoid exceedingly large test sets, we limit the number of instances in the test set to be around $215$k sentences at most (cf. Table~\ref{tab:data_stats}). The only exception is for the Multilingual and Code-Mixed subsets. The Multilingual test split was generated from the language specific test splits, and was downsampled to contain only ~471k instances. On the other hand, for the Code-Mixed subset, we sample test sentences from the language specific test splits, and replace the original entity surface forms with the surface form of the entity in another language, picked at random.

The larger size of test sets are done for two reasons: (1) to  assess the generalizability of models on unseen and complex entities; and (2) assess cross-domain adaptation performance. Table~\ref{tab:test_domains} provides a breakdown of the number of instances for the different domains across the different subtasks.

\change{Finally, the overlap of NEs between the test and train set is fairly small, with an overlap of 5.6\% across all NE classes and languages. Such a small NE overlap ensures that the test dataset is suitable for measuring NER model generalization.}

% \TK Besnik: please describe how the sets were sampled. We had some justifications for the set sizes, etc.

% Domain adaption: two domains only have 50 samples in train, etc. want to evaluate out od domain performance.

%%%% Besnik: I moved the info from section 4 to section 2 according to the recommendation from R1. She was right as the information was somehow duplicated and misplaced in S4.

\begin{table*}[h!]
\centering
\resizebox{1\textwidth}{!}{
    
    \begin{tabular}{l l r r r r r r r r r r r r r}
    \toprule
    class & split & \langid{EN} & \langid{DE} & \langid{ES} & \langid{RU} & \langid{NL} & \langid{KO} & \langid{FA} & \langid{ZH} & \langid{HI} & \langid{TR} & \langid{BN} & \langid{MULTI} & \langid{MIX}\\
\midrule
\multirow{3}{*}{\texttt{PER}} & train & 5,397 & 5,288 & 4,706 & 3,683 & 4,408 & 4,536 & 4,270 & 2,225 & 2,418 & 4,414 & 2,606 & 43,951 & 296 \\
 & dev & 290 & 296 & 247 & 192 & 212 & 267 & 201 & 129 & 133 & 231 & 144 & 2,342 & 96 \\
 & test & 55,682 & 55,757 & 51,497 & 44,687 & 49,042 & 39,237 & 35,140 & 26,382 & 25,351 & 26,876 & 24,601 & 111,346 & 19,313 \\[1ex]
\multirow{3}{*}{\texttt{LOC}} & train & 4,799 & 4,778 & 4,968 & 4,219 & 5,529 & 6,299 & 5,683 & 6,986 & 2,614 & 5,804 & 2,351 & 54,030 & 325 \\
 & dev  & 234 & 296 & 274 & 221 & 299 & 323 & 324 & 378 & 131 & 351 & 101 & 2,932 & 108 \\
 & test & 59,082 & 59,231 & 58,742 & 54,945 & 63,317 & 52,573 & 45,043 & 43,289 & 31,546 & 34,609 & 29,628 & 141,013 & 23,111 \\[1ex]
\multirow{3}{*}{\texttt{GRP}} & train & 3,571 & 3,509 & 3,226 & 2,976 & 3,306 & 3,530 & 3,199 & 713 & 2,843 & 3,568 & 2,405 & 32,846 & 248 \\
 & dev  & 190 & 160 & 168 & 151 & 163 & 183 & 164 & 26 & 148 & 167 & 118 & 1,638 & 75 \\
 & test & 41,156 & 40,689 & 38,395 & 37,621 & 39,255 & 31,423 & 27,487 & 18,983 & 22,136 & 21,951 & 19,177 & 77,328 & 16,357\\[1ex]
\multirow{3}{*}{\texttt{CORP}} & train & 3,111 & 3,083 & 2,898 & 2,817 & 2,813 & 3,313 & 2,991 & 3,805 & 2,700 & 2,761 & 2,598 & 32,890 & 294\\
 & dev  & 193 & 165 & 141 & 159 & 163 & 156 & 160 & 192 & 134 & 148 & 127 & 1,738 & 112\\
 & test & 37,435 & 37,686 & 36,769 & 35,725 & 35,998 & 30,417 & 27,091 & 25,758 & 21,713 & 21,137 & 20,066 & 75,764 & 18,478\\[1ex]
\multirow{3}{*}{\texttt{CW}} & train & 3,752 & 3,507 & 3,690 & 3,224 & 3,340 & 3,883 & 3,693 & 5,248 & 2,304 & 3,574 & 2,157 & 38,372 & 298\\
 & dev  & 176 & 189 & 192 & 168 & 182 & 196 & 207 & 282 & 113 & 190 & 120 & 2,015 & 102\\
 & test & 42,781 & 42,133 & 43,563 & 39,947 & 41,366 & 33,880 & 30,822 & 30,713 & 21,781 & 23,408 & 21,280 & 89,273 & 20,313\\[1ex]
\multirow{3}{*}{\texttt{PROD}} & train & 2,923 & 2,961 & 3,040 & 2,921 & 2,935 & 3,082 & 2,955 & 4,854 & 3,077 & 3,184 & 3,188 & 35,120 & 316\\
 & dev  & 147 & 133 & 154 & 151 & 138 & 177 & 157 & 274 & 169 & 158 & 190 & 1,848 & 117\\
 & test & 36,786 & 36,483 & 36,782 & 36,533 & 36,964 & 29,751 & 26,590 & 28,058 & 22,393 & 21,388 & 20,878 & 75,871 & 20,255\\
 \midrule
 \midrule
 \multirow{3}{*}{\#instances} & train & 15,300 & 15,300 & 15,300 & 15,300 & 15,300 & 15,300 & 15,300 & 15,300 & 15,300 & 15,300 & 15,300 &  168,300 & 1,500\\
 & dev & 800 & 800 & 800 & 800 & 800 & 800 & 800 & 800 & 800 & 800 & 800 & 8,800 & 500\\
 & test & 217,818 & 217,824 & 217,887 & 217,501 & 217,337 & 178,249 & 165,702 & 151,661 & 141,565 & 136,935 & 133,119 &  471,911 & 100,000 \\
    \bottomrule
    \end{tabular}}
    \caption{\small{\mconer dataset statistics for the different languages for the train/dev/test splits. For each NER class we show the total number of entity instances per class on the different data splits. The bottom three rows show the total number of sentences for each language.}}
    \label{tab:data_stats}
\end{table*}

\begin{table}[h!]
    \centering
    \resizebox{0.9\columnwidth}{!}{
    \begin{tabular}{l r r r | r}
    \toprule
    \emph{lang} & \small\lowner & \small\msquery & \small\msquestion & Total\\
    \midrule
     \langid{EN} &  100,000 &  100,000 & 17,818 & 217,818 \\
    \langid{DE} &  100,000 &  100,000 & 17,824 & 217,824 \\
    \langid{ES} &  100,000 &  100,000 & 17,887 & 217,887 \\
    \langid{RU} &  100,000 &  100,000 & 17,501 & 217,501 \\
    \langid{NL} &  100,000 &  100,000 & 17,337 & 217,337 \\
    \langid{KO} &  60,425 &  100,000 & 17,824 & 178,249 \\
    \langid{FA} &  48,792 &  100,000 & 16,910 & 165,702 \\
    \langid{ZH} &  33,776 &  100,000 & 17,885 & 151,661 \\
    \langid{HI} &  24,807 &  100,000 & 16,758 & 141,565 \\
    \langid{TR} &  19,581 &  100,000 & 17,354 & 136,935 \\
    \langid{BN} &  15,698 &  100,000 & 17,421 & 133,119 \\
    \langid{MULTI} &  200,000 &  200,000 & 71,911 & 471,911 \\
    \langid{MIX} &  42,168 &  15,667 & 42,165 & 100,000 \\
    \bottomrule
    \end{tabular}}
    \caption{\small{Test data statistics per domain. }}
    \label{tab:test_domains}
\end{table}

\subsection{License, Availability, and File Format}
% \textbf{License}
The dataset is released under a CC BY-SA 4.0 license, which allows adapting the data. Details about the license are available on the Creative Commons website.\footnote{\tiny{\url{https://creativecommons.org/licenses/by-sa/4.0}}}
% \noindent\textbf{Data Access}
 %\footnote{\scriptsize\url{https://registry.opendata.aws/AnonData}} 
% Additionally, it is also available via HuggingFace \TK?
% \noindent\textbf{Format}
The data is distributed using the commonly used BIO tagging scheme in \conll format~\citep{sang2003introduction}. 
The complete dataset is available for download.\footnote{\tiny\url{https://registry.opendata.aws/multiconer/}} %  through \texttt{Open Data on AWS}.

\section{Dataset Construction}
\label{sec:data_construction}

In this section, we provide a detailed description of the methods used to generate our dataset.

\subsection{Entity Gazetteer Data}
\label{sec:gaz_data}

We require a large, multilingual, and reliable source of known entities for generating our dataset. To this end we leverage the Wikidata to obtain entity information. Instead of using all entities in the KB, or collecting entities from web sources \citep{daniel2018cogcompnlp},
we instead focus on entities that map to our taxonomy.

We map Wikidata entities to our NER taxonomy (\S\ref{sec:taxonomy}). This is done by traversing Wikidata's class and instance relations, and manually mapping them to our NER classes, \eg Wikidata's \classname{human} class maps to \classname{PER} in our taxonomy, \classname{song} to \classname{CW}, and so on. Alternative names (aliases) for entities are included.
%Gazetteer statistics are listed in \Cref{sec:app-gaz}.
%We extracted 1.67 million entities that were mapped to our classes.
The distribution of these entities is shown in \Cref{tab:gaz_data} in \S\ref{subsec:gazetteer_stats}.

% \begin{table}[t!]
% \resizebox{\columnwidth}{!}{
%     %\begin{center}
%     \begin{tabular}{l r r r r r r}
%     \toprule
%     \emph{lang} & \texttt{PER} & \texttt{LOC} & \texttt{CORP} & \texttt{GRP} & \texttt{PROD} & \texttt{CW}\\
%     \midrule
%     \langid{BN} & 42,970 & 31,336 & 1,072 & 8,691 & 990 & 12,152 \\
%     \langid{ZH} & 388,910 & 346,879 & 30,323 & 64,031 & 15,919 & 120,831 \\
%     \langid{NL} & 1,321,741 & 738,609 & 27,589 & 79,566 & 21,105 & 204,130 \\
%     \langid{EN} & 1,797,558 & 1,117,951 & 72,105 & 227,822 & 67,113 & 490,523 \\ 
%     \langid{FA} & 224,265 & 233,962 & 8,641 & 14,346 & 11,802 & 60,857 \\ 
%     \langid{DE} & 1,308,532 & 533,551 & 42,321 & 99,468 & 38,735 & 219,801 \\ 
%     \langid{HI} & 22,279 & 18,480 & 1,160 & 2,382 & 1,044 & 7,826 \\ 
%     \langid{KO} & 148,367 & 72,153 & 9,625 & 23,209 & 8,385 & 55,624 \\ 
%     \langid{RU} & 984,093 & 495,059 & 21,609 & 68,834 & 21,571 & 148,003 \\ 
%     \langid{ES} & 1,389,698 & 480,310 & 29,465 & 113,197 & 25,658 & 228,369 \\ 
%     \langid{TR} & 171,133 & 141,225 & 6,099 & 19,388 & 6,718 & 43,029\\
%     \bottomrule
%     \end{tabular}
%     %\end{center}
%     }
% \caption{\small{Gazetteer entity statistics per class for our target languages.}}
% \label{tab:gaz_data}
% \end{table}

\subsection{Wiki Sentences}\label{subsec:wiki_sentences}

\begin{figure*}
    \centering
    \pdftooltip{\includegraphics[width=0.7\linewidth]{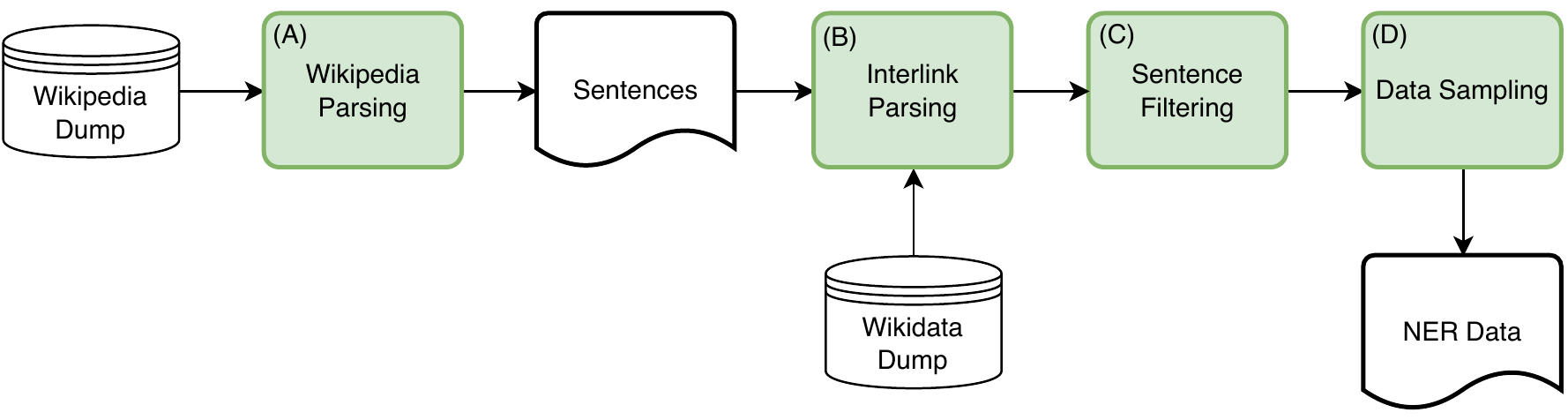}}{TOOLTIP.}
\caption{An overview of the different steps involved in extracting the \mconer data from Wikipedia dumps.}
\label{fig:lowner_dfd}
\end{figure*}

\lowner, the Wiki sentences component of \mconer is obtained by parsing Wikipedia articles into sentences, and selecting suitable candidates.
\Cref{fig:lowner_dfd} shows a diagram of the basic data processing steps, which are described below.

This process is performed for the following languages:  \langid{NL}, \langid{EN}, \langid{FA}, \langid{KO}, \langid{RU}, \langid{ES}, \langid{TR}.
For the other languages (\langid{BN}, \langid{ZH}, \langid{DE}, \langid{HI}), we apply Machine Translation to obtain the data, as described in \S\ref{sec:translation}.

\paragraph{Wikipedia Parsing (A)}
We start by downloading the complete Wikipedia dumps for our target languages.\footnote{Dumps are available here: \url{https://dumps.wikimedia.org/backup-index.html}}
The files are parsed to first extract individual articles, which are then each parsed to remove markup and extract individual sentences.
This process yields a set of sentences\footnote{\eg > 180 million sentences for English.} with the interlinks intact, along with the IDs of the original article they were extracted from.

\paragraph{Interlink Parsing (B)}
In the next step, we parse the sentences to identify interlinks (links to other Wikipedia articles). We then map the interlinks in each sentence to an entity in the Wikidata KB. This mapping is provided in the KB, which links entities to their Wikipedia page names, which can be used to map linked pages to an entity ID.
The identified entities are finally resolved to our NER taxonomy (using the same approach that was described in \S\ref{sec:gaz_data}).
Some interlinks point to inexisting Wikipedia articles, or the linked Wikipedia article cannot be joined to a corresponding Wikidata entry. We mark such cases as unresolvable.

\paragraph{Sentence Filtering (C)}
Next, we filtered sentences using several strategies and heuristics.

\begin{itemize}[leftmargin=*]
    \itemsep0em
    \item Length filtering: short sentences (< 28 characters) and long ones (> 180 characters) are removed.
    
    \item Interlink filtering: sentences without interlinks, or with unresolvable links, are dropped. Sentences with interlinks that did not map to our taxonomy are dropped.
    
    \item Capitalization heuristic filtering: for languages that capitalize proper nouns or entities, a heuristic is used to filter out sentences that contain capitalized tokens that are not part of an interlink. This removes sentences containing potential nouns that cannot be tagged as entities by our method since they are not linked to a known entity whose type can be determined.
\end{itemize}

%Taking advantage of Wikipedia's well-formed text, we applied a Regex-based NER method to identify sentences containing named entities that were not linked, and removed them.
% Additionally we also removed any sentence where the links could not be resolved to Wikidata entities.

This filtering process removes long and high-context sentences that contain references to many entities.  This step discards over 90\% of the sentences retrieved in the prior steps, \eg resulting in approx. 14 million candidate sentences for \langid{EN}. \change{Finally, to assess the NER label quality, for a small random sample of 400 sentences, we assessed the accuracy of NER gold labels, which was was measured at 94\% accuracy for \langid{EN}-\lowner.}

This process is very effective at yielding short, low-context sentences. Example English sentences are shown in \Cref{tab:examples_wiki}. The sentences contain some context, but they are much shorter than the average Wikipedia sentence, and usually only contain a single entity, making them more aligned with the challenges listed in \Cref{sec:ner_challenges}.

\begin{table}[h!]
\centering
\resizebox{1\columnwidth}{!}{
\small{
\begin{tabular}{p{10cm}}
\toprule
\example{The design is considered a forerunner to the modern [food processor].} \\
\example{The regional capital is [Oranjestad, Sint Eustatius].}\\
\example{The most frequently claimed loss was an [iPad].}\\
% \example{A [Macintosh] version was released in 1994.}\\
\example{An [HP TouchPad] was prominently displayed in an episode of the sixth season.}\\
\example{The incumbent island governor is [Jonathan G. A. Johnson].}\\
\example{A revised edition of the book was released in 2017 as an [Amazon Kindle] book.}\\
\bottomrule
\end{tabular}}}
\caption{Sample sentences extracted from Wikipedia. Resolved entities are in brackets.}
\label{tab:examples_wiki}
\end{table}

\paragraph{Data Sampling (D)}

We downsample the collected data to construct a smaller subset.  Given that some of the NE classes are more prevalent (e.g. \classname{PER} and \classname{LOC}, account for more than 60\% of named entities) , similar to stratified sampling, we sample at NE class level, with the only difference, that the number of instances per class is fixed. In this way, we create a dataset that has more uniform representation of the different NE classes. Furthermore, retaining all sentences is impractical, given the total amount of data.
% that the extracted data for some of the subsets are quite large.

% we apply stratified sampling to create a smaller subset of the data that is also equally representative of all of our entity classes.

% We did not want to use the entire set of sentences as it would be extremely large. We also did not want to take a simple random subsample as the distribution of entities in Wikipedia is imbalanced.

% \TK We need to add more details. Besnik can you please add some more information about the original entity distribution and how it is imbalanced, and the stratified sampling that was applied. Please also verify that the data is indeed lowercased.

Finally, we lowercase all the selected sentences to increase the difficulty of the NER task. The final stats per subset and split are shown in Table~\ref{tab:data_stats}.

\subsection{Questions and Search Queries}

We apply a template-based process to generate data in the Questions and Search Query domains. This involves two broad steps: template extraction, and template slotting.

\begin{figure*}[ht!]
    \centering
    \pdftooltip{\includegraphics[width=0.8\linewidth]{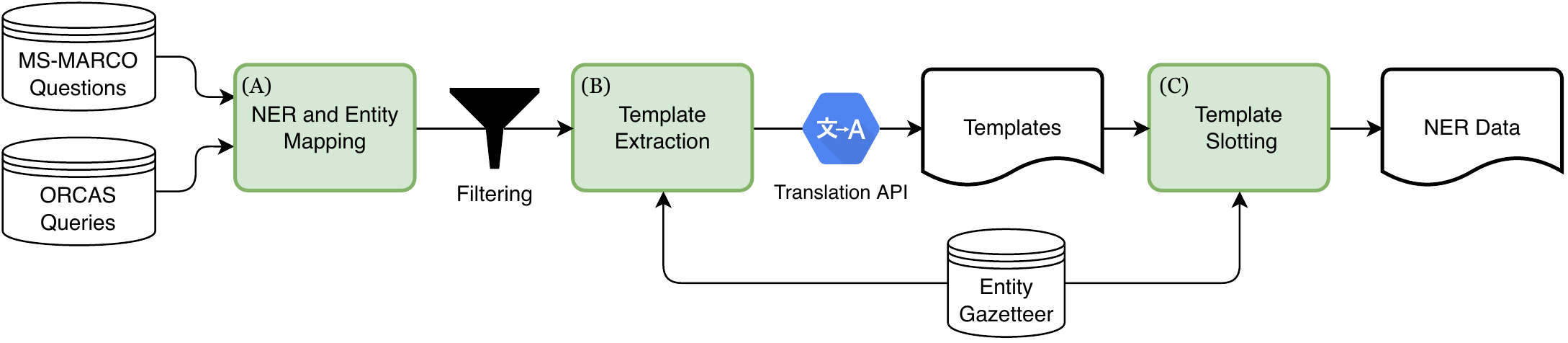}}{TOOLTIP.}
\caption{\small{An overview of the data processing steps in our template-based approach to generating the NER data for the \msquestion and \msquery domains.}}
\label{fig:slotting_dfd}
\end{figure*}

The same steps are applied to two data sources to generate the NER datasets.
This process is visualized in \Cref{fig:slotting_dfd}, and the individual steps are detailed below.

\begin{table}
\resizebox{1.0\columnwidth}{!}{
\begin{tabular}{l l}
\toprule
\msquestion & \msquery \\
\midrule
\example{average retail price of <PROD>} & \example{<CW> imdb}\\
\example{where was <CW> filmed} & \example{best hotels <LOC>}\\
\example{how many miles from <LOC> to <LOC>} & \example{<PER> parents}\\
\example{how many kids does <PER> have} & \example{<PROD> price} \\
\example{when did <GRP> start} & \example{<GRP> website} \\
\example{when will <CORP> report earnings} & \example{<CORP> customer service} \\
\bottomrule
\end{tabular}}
\caption{Sample templates used to generate data for the \msquestion and \msquery domain. Slots are in angle brackets.}
\label{tab:examples_msq}
\end{table}

\paragraph{Running Named Entity Recognition (A)}
Similar to the work of \citet{wu2020tempura}, we aim to templatize the input questions and search queries by first applying NER to extract entities, which are then mapped to our taxonomy.

Specifically, we apply the spaCy NER pipeline\footnote{We use the \texttt{en\_core\_we\_lg} pre-trained pipeline in spaCy where the NER model is trained using OntoNotes 5.} to identify entities. While this pre-trained NER system cannot correctly identify all the entities in the data, it does identify many correct ones. 
This process yields a sufficient amount of data for us to extract common patterns from the original input.
%the most frequent templates are reliable.
%
Recognized entities are then mapped to entries in our gazetteer via string matching.
Input texts that have entities that could not be mapped, or have no recognized entities are then filtered out.
This process yields a set of sentences, with recognized entities that exist in our gazetteer.

% \begin{table}
% \centering\small{
% \begin{tabular}{l}
% \toprule
% \example{<CW> imdb} \\
% \example{best hotels <LOC>} \\
% \example{<PER> parents} \\
% \example{<PROD> price} \\
% \example{<GRP> website} \\
% \example{<CORP> customer service} \\
% \bottomrule
% \end{tabular}}
% \caption{Sample search query templates use to generate data for the \msquery domain. Slots are in angle brackets.}
% \label{tab:examples_orcas}
% \end{table}

\paragraph{Template Extraction (B)}
Next, we replace identified entities with their types to create templates, \eg \exampleq{when did [iphone] come out} is transformed to \exampleq{when did <PROD> come out}.
The templates are then grouped together in order to merge all inputs having the same template, and sorted by frequency.

To minimize noise in the data, we apply frequency-based filtering, and only templates appearing $\smash{>=5}$ times are included.
This process results in $3,445$ unique question templates, and $97,324$ unique search query templates.
There are a wide range of question shapes and entity types. Examples are listed in \Cref{tab:examples_msq}.

Since these templates are all in English, we apply Machine Translation to translate them into the other 10 languages of our dataset.

\paragraph{Template Slotting (C)}

In the last step we generate the NER data by slotting the templates with random entities from the Wikipedia KB with the same class. For example,  \exampleq{when did <PROD> come out} can be slotted as \exampleq{when did [xbox 360] come out} or \exampleq{when did [Sony Alpha DSLR-A77 II] come out}.

To maintain the same relative distribution as the original data, each template is slotted the same number of times it appeared (\ie the template frequency) using different entities.
Templates are slotted with entities from the same language, \ie a \langid{DE} template is slotted with \langid{DE} entities.
The slotted texts are lowercased to simulate the low-context challenges outlined in \S\ref{sec:ner_challenges}, which increases the difficulty of the task.
This yields two domains of \mconer: \msquestion and \msquery.

\begin{table}
\centering
\resizebox{1\columnwidth}{!}{
\small{
\begin{tabular}{p{11cm}}
\toprule
\example{EN: average cost of living in <p translate=no> <LOC> </p>} \\
\example{ZH: \zh{<p translate=no> <LOC> </p>的平均生活成本}} \\
\example{DE: durchschnittliche Lebenshaltungskosten in <p translate=no> <LOC> </p>} \\
\example{HI: {\dn rhne ki aust laa} <p translate=no> <LOC> </p>} \\
% \example{BN: {জীবনযাত্রার গড় খরচ <p translate=no> <LOC> </p> }} \\
\bottomrule
\end{tabular}}}
\caption{Examples of template translations.}
\label{tab:translation_example}
\end{table}

\subsection{Dataset Translation}
\label{sec:translation}
We apply automatic translation to generate two portions of our data.
\lowner sentences for four languages (Bangla, Chinese, German, Hindi) are translated from English Wiki sentences.
The NER templates used for \msquestion and \msquery are also translated from the English templates. We do not translate any of our gazetteer entities.

We use the Google Translation API\footnote{\url{https://cloud.google.com/translate}} to perform our translations. 
The input texts may contain known entity spans or slots. 
%To make sure the entity placeholder remain the same, we use a no translation html tag \texttt{``<p translate=no> xxx <\p>''} in the translation.
To prevent these spans from being translated, we leveraged the \texttt{notranslate} attribute to mark these spans and prevent them from being translated.
\Cref{tab:translation_example} shows examples of template translations.

\change{The translation quality of \lowner, \msquery and \msquestion in the different languages such as  German, Chinese, Bangla, and Hindi is very high, with over 90\% translation accuracy (i.e., accuracy as measured by human annotators in terms of the translated sentence retaining the semantic meaning and as well have a correct syntactic structure in the target language).}

\subsection{Code-mixed Data Generation}

We generate code-mixed data by sampling instances from the respective languages, and replacing the NE surface forms from the \emph{source} language to a \emph{target} language, chosen at random from any of the languages in Table~\ref{tab:languages}. This results in a dataset, where the instances contain up to two languages, where the non-NE tokens are in a language that is different from the NE tokens. Note that, in some cases, some of the NE surface forms may remain in the source language if we do not possess the NE's surface form in one of the other languages from Table~\ref{tab:languages}.\footnote{For \langid{ZH}, the tokenization is done at the character level.}

\begin{table*}
    \centering
    \resizebox{1.0\textwidth}{!}{
    \begin{tabular}{l l l | l l | l l | l l | l l | l l | l l | l l | l l}
    \toprule
         & \multicolumn{2}{c}{\texttt{PER}} & \multicolumn{2}{c}{\texttt{LOC}} & \multicolumn{2}{c}{\texttt{GRP}} & \multicolumn{2}{c}{\texttt{CORP}} & \multicolumn{2}{c}{\texttt{CW}} & \multicolumn{2}{c}{\texttt{PROD}} & \multicolumn{2}{c}{Micro F1} & \multicolumn{2}{c}{Macro F1} & \multicolumn{2}{c}{MD}\\
         \midrule
         & \xlmr & \gemnet & \xlmr & \gemnet & \xlmr & \gemnet & \xlmr & \gemnet & \xlmr & \gemnet & \xlmr & \gemnet & \xlmr & \gemnet & \xlmr & \gemnet & \xlmr & \gemnet\\
         \midrule
\langid{EN} & 0.807 & 0.939 & 0.664 & 0.848 & 0.599 & 0.876 & 0.567 & 0.889 & 0.474 & 0.806 & 0.563 & 0.873 & 0.627 & 0.873 & 0.612 & 0.872 & 0.731 & 0.892\\
\langid{DE} & 0.797 & 0.968 & 0.679 & 0.921 & 0.591 & 0.940 & 0.588 & 0.951 & 0.516 & 0.897 & 0.633 & 0.940 & 0.646 & 0.936 & 0.634 & 0.936 & 0.767 & 0.951\\
\langid{ES} & 0.750 & 0.941 & 0.589 & 0.854 & 0.531 & 0.884 & 0.564 & 0.893 & 0.496 & 0.811 & 0.515 & 0.840 & 0.587 & 0.872 & 0.574 & 0.870 & 0.689 & 0.888\\
\langid{RU} & 0.666 & 0.839 & 0.632 & 0.780 & 0.539 & 0.818 & 0.600 & 0.870 & 0.534 & 0.803 & 0.576 & 0.805 & 0.597 & 0.817 & 0.591 & 0.819 & 0.699 & 0.834\\
\langid{NL} & 0.766 & 0.949 & 0.658 & 0.889 & 0.586 & 0.893 & 0.599 & 0.905 & 0.514 & 0.854 & 0.574 & 0.871 & 0.626 & 0.895 & 0.616 & 0.894 & 0.731 & 0.911\\
\langid{KO} & 0.595 & 0.900 & 0.650 & 0.865 & 0.513 & 0.910 & 0.560 & 0.923 & 0.439 & 0.846 & 0.517 & 0.905 & 0.558 & 0.888 & 0.546 & 0.891 & 0.666 & 0.896\\
\langid{FA} & 0.634 & 0.870 & 0.588 & 0.792 & 0.573 & 0.867 & 0.473 & 0.805 & 0.362 & 0.688 & 0.480 & 0.797 & 0.523 & 0.801 & 0.518 & 0.803 & 0.638 & 0.823\\
\langid{TR} & 0.549 & 0.894 & 0.497 & 0.860 & 0.404 & 0.896 & 0.480 & 0.897 & 0.374 & 0.849 & 0.441 & 0.914 & 0.468 & 0.883 & 0.457 & 0.885 & 0.614 & 0.893\\
\langid{ZH} & 0.532 & 0.884 & 0.627 & 0.889 & 0.371 & 0.866 & 0.552 & 0.902 & 0.434 & 0.818 & 0.552 & 0.861 & 0.531 & 0.870 & 0.511 & 0.870 & 0.664 & 0.895\\
\langid{HI} & 0.578 & 0.883 & 0.536 & 0.846 & 0.485 & 0.869 & 0.502 & 0.851 & 0.298 & 0.760 & 0.418 & 0.839 & 0.478 & 0.843 & 0.469 & 0.841 & 0.640 & 0.877\\
\langid{BN} & 0.529 & 0.895 & 0.420 & 0.850 & 0.322 & 0.883 & 0.428 & 0.889 & 0.243 & 0.747 & 0.406 & 0.865 & 0.397 & 0.856 & 0.391 & 0.855 & 0.570 & 0.888\\

\langid{MULTI} & 0.679 & 0.810 & 0.556 & 0.743 & 0.496 & 0.721 & 0.563 & 0.746 & 0.428 & 0.644 & 0.523 & 0.697 & 0.550 & 0.732 & 0.541 & 0.727 & 0.674 & 0.810\\
\langid{MIX} & 0.709 & 0.835 & 0.621 & 0.765 & 0.532 & 0.714 & 0.581 & 0.748 & 0.481 & 0.604 & 0.560 & 0.735 & 0.585 & 0.731 & 0.581 & 0.733 & 0.752 & 0.847\\

\midrule\midrule
\langid{Avg.} & 0.661 & 0.893 & 0.594 & 0.839 & 0.503 & 0.857 & 0.543 & 0.867 & 0.430 & 0.779 & 0.520 & 0.842 & 0.552 & 0.846 & 0.542 & 0.846 & 0.680 & 0.877\\

\bottomrule
    \end{tabular}}
    \caption{XLM-RoBERTa (B) baseline and  {GEMNET} (GM) results as measured by the F1 score for the different NER tags. In the last three columns are shown the \emph{micro}, \emph{macro}, and \emph{mention detection} -- MD F1 performance. }
    \label{tab:baseline_results}
\end{table*}

\section{NER Model Performance}
To evaluate whether our new dataset poses real-world challenges (cf. \S\ref{sec:ner_challenges}), we train and test two existing NER systems:
(1) XLM-RoBERTa~\cite{DBLP:conf/acl/ConneauKGCWGGOZ20}, a large multilingual Transformer model; and (2) GEMNET~\cite{DBLP:conf/naacl/MengFRM21,fetahu2021gazetteer,DBLP:conf/naacl/FetahuFRM22}, a state of the art model that integrates gazetteers into XLM-RoBERTa. 

\subsection{Evaluation Metrics}

We evaluate the different NER models using standard performance metrics, such as P/R/F1. We measure the performance at the class level, where we distinguish between \emph{micro/macro} averages. The difference between \emph{micro} and \emph{macro} average P/R/F1, is that for unbalanced distribution \emph{micro} performance is skewed towards the more prominent NE classes.
Additionally, we consider \emph{Mention Detection} (MD), which corresponds to the ability of models to detect NE boundaries, without taking into consideration their actual NER class.

% Due to space constrains, in Table~\ref{tab:baseline_results}, we only report the F1 scores per class, micro/macro F1, and the MD scores.

\subsection{Results}
% \subsubsection{Overall Evaluation Results}

Table~\ref{tab:baseline_results} shows the results obtained for both XLM-RoBERTa (baseline, denoted as B), and a state of the art model, GEMNET (denoted as GM).
The results are shown only for the F1 score achieved on the individual NER classes, and finally the micro, macro F1 and MD scores are shown. Table~\ref{tab:domain_results} in \S\ref{subsec:domain_results} shows a detailed performance for each subtask and domain.

\textbf{Baseline.} For the baseline approach, we simply fine-tune XLM-RoBERTa on the language specific training data, and test on the corresponding test splits. We note that overall, across all subsets, the baseline achieves the highest performance of micro-F1=0.646 for DE, and lowest of micro-F1=0.397 for BN. This result is expected, given that the test data contains highly complex entities that are out-of-domain, and additionally are not seen in the training data. 

\textbf{GEMNET.} The state of the art approach, GEMNET, makes use of external gazetteers, constructed from Wikidata for the task of NER. For each token GEMNET computes two representations: (i) textual representation based on XLM-RoBERTa, and (ii) gazetteer encoding, which uses a gazetteer to map to tokens to gazetteer entries, which correspondingly maps them to their NER tags. The two representations are combined using a Mixture-of-Experts (MoE) gating mechanism~\cite{DBLP:conf/iclr/ShazeerMMDLHD17}, which allows the model depending on the context to either assign higher weight to its textual representation or the gazetteer computed representation.

GEMNET provides a highly significant improvement over the baseline, with an average improvement of micro-F1=30\%. The highest improvement is shown for languages that are considered to be low-resource, such as \langid{TR} with micro-F1=+41.5\%, and \langid{KO} with micro-F1=+33\%.

The obtained results in Table~\ref{tab:baseline_results} show that GEMNET is highly flexible in detecting unseen entities during the training phase. Depending on its gazetteer coverage,
% \footnote{It is assumed that when GEMNET is applied on an out-of-domain corpus, target named entities are present in its gazetteer, allowing GEMNET to obtain highly optimal performance without further fine-tuning.} 
if a named entity is matched by its gazetteers, this will allow GEMNET to correctly identify the named entity. In more detail, internally, GEMNET dynamically weighs both representation of a token (i.e., textual and gazetteer representations), to correctly determine the correct tag for a token. Note that, the gazetteers may contain noisy labels for a named entity (e.g. \example{``Bank of America''} can match to \classname{CORP} and \classname{LOC}), hence, GEMNET needs to additionally leverage the token context to determine the correct tag.  

% \subsubsection{Cross-Domain Evaluation Results}

% Table~\ref{tab:domain_results} shows for the different subtasks, the cross-domain evaluation results for the baseline and GEMNET approaches. We note that in all cases the GEMNET approach shows strong gains in terms of macro-F1 score across all subtasks. This is especially the case for the domains MSQ and ORCAS, where the models contain very little knowledge about these domains\footnote{The AnonData training set for each of the subtasks, contains 50 instances from the MSQ and ORCAS domains}, hence, showing the generalizability of models in out-of-domain scenarios. 

% Finally, we note that in the case of the LOWNER domain, which is an in-domain evaluation scenario\footnote{The AnonData training set consists nearly of only LOWNER instances.}, in terms of MD, the gap between the baseline and GEMNET approach shrinks. For the \langid{Multi}, the gap is only 4.1\%. This shows that the baseline models for in-domain scenarios is able to correctly identify entity boundaries, even though its NER accuracy may not be optimal. For instance, for \langid{Multi} the gap in terms of macro-F1 is 12.7\%. showing, that models that leverage external knowledge like GEMNET, are more accurate in terms of NER accuracy and as well have higher coverage in spotting entity boundaries.

\section{Conclusions and Future Work}

We presented \mconer, a new large-scale dataset that represents a number of current challenges in NER.
Results obtained on our data showed that our dataset is challenging. %\TK (summarize results).
A XLMR based model achieves only approx. 50\% F1 in average while GEMNET improves F1 performance by more than 30\% using gazetteers. 

These results demonstrate that \mconer represents challenging scenarios where even large pre-trained language models fail to achieve good performance without external resources.
It is our hope that this resource will help further research for building better NER systems. This dataset can serve as a good benchmark for evaluating different methods of infusing external entity knowledge into language models.

The extension of \mconer to additional languages is the most straightforward direction for future work.
The addition of completely new domains is something we will also consider, along with the the expansion of the existing domains to include additional topics and templates. 

% \balance 
\bibliography{tacl2021}
\bibliographystyle{acl_natbib}

\clearpage
\newpage
\appendix

\section{Appendix}\label{sec:appendix}
\subsection{Gazetteer Statistics}\label{subsec:gazetteer_stats}
Table~\ref{tab:gaz_data} shows the number of entries for NE class and language. The entries are extracted from Wikidata.
\begin{table}[h!]
\resizebox{\columnwidth}{!}{
    %\begin{center}
    \begin{tabular}{l r r r r r r}
    \toprule
    \emph{lang} & \texttt{PER} & \texttt{LOC} & \texttt{CORP} & \texttt{GRP} & \texttt{PROD} & \texttt{CW}\\
    \midrule
    \langid{BN} & 42,970 & 31,336 & 1,072 & 8,691 & 990 & 12,152 \\
    \langid{ZH} & 388,910 & 346,879 & 30,323 & 64,031 & 15,919 & 120,831 \\
    \langid{NL} & 1,321,741 & 738,609 & 27,589 & 79,566 & 21,105 & 204,130 \\
    \langid{EN} & 1,797,558 & 1,117,951 & 72,105 & 227,822 & 67,113 & 490,523 \\ 
    \langid{FA} & 224,265 & 233,962 & 8,641 & 14,346 & 11,802 & 60,857 \\ 
    \langid{DE} & 1,308,532 & 533,551 & 42,321 & 99,468 & 38,735 & 219,801 \\ 
    \langid{HI} & 22,279 & 18,480 & 1,160 & 2,382 & 1,044 & 7,826 \\ 
    \langid{KO} & 148,367 & 72,153 & 9,625 & 23,209 & 8,385 & 55,624 \\ 
    \langid{RU} & 984,093 & 495,059 & 21,609 & 68,834 & 21,571 & 148,003 \\ 
    \langid{ES} & 1,389,698 & 480,310 & 29,465 & 113,197 & 25,658 & 228,369 \\ 
    \langid{TR} & 171,133 & 141,225 & 6,099 & 19,388 & 6,718 & 43,029\\
    \bottomrule
    \end{tabular}
    %\end{center}
    }
\caption{\small{Gazetteer entity statistics per class for our target languages.}}
\label{tab:gaz_data}
\end{table}

\subsection{Cross-Domain Evaluation Results}\label{subsec:domain_results}

Table~\ref{tab:domain_results} shows for the different subtasks, the cross-domain evaluation results for the baseline and GEMNET approaches. We note that in all cases the GEMNET approach shows strong gains in terms of macro-F1 score across all subtasks. This is especially the case for the domains  \msquestion and \msquery, where the models contain very little knowledge about these domains\footnote{The MultiCoNER training set for each of the subtasks, contains 50 instances from the \msquestion and \msquery domains}, hence, showing the generalizability of models in out-of-domain scenarios. 

Finally, we note that in the case of the \lowner domain, which is an in-domain evaluation scenario\footnote{The MultiCoNER training set consists nearly of only \lowner instances.}, in terms of MD, the gap between the baseline and GEMNET approach shrinks. For the \langid{Multi}, the gap is only 4.1\%. This shows that the baseline models for in-domain scenarios is able to correctly identify entity boundaries, even though its NER accuracy may not be optimal. For instance, for \langid{Multi} the gap in terms of macro-F1 is 12.7\%. showing, that models that leverage external knowledge like GEMNET, are more accurate in terms of NER accuracy and as well have higher coverage in spotting entity boundaries.

\begin{table*}[h!]
    \centering
    \resizebox{1.0\textwidth}{!}{
    \begin{tabular}{l l l | l l | l l | l l | l l | l l | l l | l l | l l}
    \toprule
         & \multicolumn{2}{c}{\texttt{PER}} & \multicolumn{2}{c}{\texttt{LOC}} & \multicolumn{2}{c}{\texttt{GRP}} & \multicolumn{2}{c}{\texttt{CORP}} & \multicolumn{2}{c}{\texttt{CW}} & \multicolumn{2}{c}{\texttt{PROD}} & \multicolumn{2}{c}{Micro F1} & \multicolumn{2}{c}{Macro F1} & \multicolumn{2}{c}{MD}\\
         \midrule
         \midrule
         \multicolumn{18}{c}{\emph{Domain} -- \lowner}\\
         \midrule
         & \xlmr & \gemnet & \xlmr & \gemnet & \xlmr & \gemnet & \xlmr & \gemnet & \xlmr & \gemnet & \xlmr & \gemnet & \xlmr & \gemnet & \xlmr & \gemnet & \xlmr & \gemnet\\
         \midrule
\langid{EN} & 0.921 & 0.971  & 0.855 & 0.938  & 0.766 & 0.925  & 0.756 & 0.939  & 0.681 & 0.862  & 0.656 & 0.849  & 0.789 & 0.920  & 0.773 & 0.914  & 0.851 & 0.932 \\
\langid{DE} & 0.913 & 0.978  & 0.871 & 0.957  & 0.781 & 0.948  & 0.776 & 0.952  & 0.706 & 0.913  & 0.772 & 0.921  & 0.816 & 0.949  & 0.803 & 0.945  & 0.903 & 0.965 \\
\langid{ES} & 0.897 & 0.944  & 0.797 & 0.866  & 0.725 & 0.871  & 0.792 & 0.910  & 0.667 & 0.802  & 0.627 & 0.761  & 0.759 & 0.864  & 0.751 & 0.859  & 0.821 & 0.883 \\
\langid{RU} & 0.734 & 0.794  & 0.702 & 0.757  & 0.695 & 0.794  & 0.745 & 0.855  & 0.687 & 0.793  & 0.647 & 0.753  & 0.702 & 0.788  & 0.702 & 0.791  & 0.752 & 0.809 \\
\langid{NL} & 0.904 & 0.949  & 0.878 & 0.926  & 0.797 & 0.900  & 0.801 & 0.898  & 0.732 & 0.840  & 0.715 & 0.810  & 0.816 & 0.894  & 0.805 & 0.887  & 0.871 & 0.913 \\
\langid{KO} & 0.774 & 0.896  & 0.817 & 0.885  & 0.734 & 0.882  & 0.760 & 0.910  & 0.710 & 0.850  & 0.714 & 0.852  & 0.761 & 0.880  & 0.751 & 0.879  & 0.810 & 0.890 \\
\langid{TR} & 0.813 & 0.897  & 0.825 & 0.875  & 0.807 & 0.906  & 0.798 & 0.906  & 0.684 & 0.831  & 0.640 & 0.805  & 0.768 & 0.871  & 0.761 & 0.870  & 0.818 & 0.884 \\
\langid{ZH} & 0.869 & 0.917  & 0.855 & 0.924  & 0.534 & 0.795  & 0.740 & 0.859  & 0.659 & 0.816  & 0.655 & 0.834  & 0.743 & 0.868  & 0.719 & 0.858  & 0.811 & 0.901 \\
\langid{HI} & 0.792 & 0.837  & 0.732 & 0.813  & 0.710 & 0.757  & 0.651 & 0.713  & 0.487 & 0.578  & 0.524 & 0.634  & 0.649 & 0.722  & 0.649 & 0.722  & 0.765 & 0.813 \\
\langid{BN} & 0.822 & 0.853  & 0.769 & 0.823  & 0.701 & 0.780  & 0.666 & 0.725  & 0.569 & 0.663  & 0.576 & 0.679  & 0.680 & 0.752  & 0.684 & 0.754  & 0.814 & 0.859 \\
\langid{MULTI} & 0.855 & 0.916  & 0.808 & 0.882  & 0.717 & 0.868  & 0.733 & 0.884  & 0.664 & 0.825  & 0.648 & 0.808  & 0.741 & 0.868  & 0.737 & 0.864  & 0.852 & 0.893 \\
\langid{MIX} & 0.855 & 0.862  & 0.808 & 0.809  & 0.717 & 0.737  & 0.733 & 0.757  & 0.664 & 0.616  & 0.648 & 0.719  & 0.741 & 0.749  & 0.737 & 0.750  & 0.852 & 0.850 \\

\midrule\midrule
\multicolumn{18}{c}{\emph{Domain} -- \msquestion}\\
\midrule
\langid{EN} & 0.781 & 0.921  & 0.613 & 0.823  & 0.366 & 0.788  & 0.408 & 0.801  & 0.348 & 0.795  & 0.355 & 0.852  & 0.598 & 0.842  & 0.479 & 0.830  & 0.733 & 0.860 \\
\langid{DE} & 0.758 & 0.984  & 0.708 & 0.958  & 0.317 & 0.939  & 0.397 & 0.964  & 0.415 & 0.909  & 0.346 & 0.948  & 0.643 & 0.959  & 0.490 & 0.950  & 0.783 & 0.970 \\
\langid{ES} & 0.700 & 0.979  & 0.526 & 0.879  & 0.216 & 0.857  & 0.403 & 0.924  & 0.388 & 0.856  & 0.335 & 0.885  & 0.529 & 0.901  & 0.428 & 0.897  & 0.643 & 0.912 \\
\langid{RU} & 0.692 & 0.961  & 0.652 & 0.864  & 0.317 & 0.904  & 0.436 & 0.842  & 0.435 & 0.915  & 0.280 & 0.806  & 0.601 & 0.891  & 0.469 & 0.882  & 0.726 & 0.904 \\
\langid{NL} & 0.745 & 0.980  & 0.511 & 0.895  & 0.273 & 0.831  & 0.450 & 0.947  & 0.436 & 0.922  & 0.342 & 0.937  & 0.546 & 0.919  & 0.460 & 0.919  & 0.680 & 0.932 \\
\langid{KO} & 0.547 & 0.864  & 0.644 & 0.917  & 0.255 & 0.903  & 0.370 & 0.947  & 0.288 & 0.907  & 0.235 & 0.924  & 0.531 & 0.904  & 0.390 & 0.910  & 0.669 & 0.908 \\
\langid{FA} & 0.674 & 0.914  & 0.512 & 0.789  & 0.533 & 0.829  & 0.413 & 0.805  & 0.236 & 0.740  & 0.331 & 0.762  & 0.499 & 0.814  & 0.450 & 0.807  & 0.615 & 0.840 \\
\langid{TR} & 0.597 & 0.881  & 0.568 & 0.905  & 0.246 & 0.875  & 0.389 & 0.956  & 0.357 & 0.890  & 0.211 & 0.873  & 0.517 & 0.897  & 0.395 & 0.897  & 0.647 & 0.908 \\
\langid{ZH} & 0.534 & 0.947  & 0.709 & 0.957  & 0.401 & 0.907  & 0.432 & 0.941  & 0.390 & 0.920  & 0.283 & 0.843  & 0.588 & 0.945  & 0.458 & 0.919  & 0.743 & 0.961 \\
\langid{HI} & 0.725 & 0.955  & 0.715 & 0.925  & 0.464 & 0.925  & 0.572 & 0.929  & 0.360 & 0.899  & 0.280 & 0.827  & 0.656 & 0.928  & 0.519 & 0.910  & 0.802 & 0.952 \\
\langid{BN} & 0.589 & 0.950  & 0.468 & 0.879  & 0.000 & 0.000  & 0.433 & 0.942  & 0.298 & 0.821  & 0.239 & 0.793  & 0.465 & 0.891  & 0.338 & 0.731  & 0.643 & 0.915 \\
\langid{MULTI} & 0.628 & 0.775  & 0.571 & 0.751  & 0.271 & 0.503  & 0.401 & 0.602  & 0.323 & 0.539  & 0.306 & 0.463  & 0.531 & 0.712  & 0.417 & 0.605  & 0.688 & 0.817 \\
\langid{MIX} & 0.629 & 0.857  & 0.477 & 0.764  & 0.445 & 0.733  & 0.521 & 0.786  & 0.349 & 0.666  & 0.532 & 0.777  & 0.496 & 0.763  & 0.492 & 0.764  & 0.738 & 0.891 \\

         \midrule\midrule
\multicolumn{18}{c}{\emph{Domain} -- \msquery}\\
\midrule

\langid{EN} & 0.588 & 0.886  & 0.340 & 0.719  & 0.313 & 0.811  & 0.342 & 0.834  & 0.191 & 0.736  & 0.430 & 0.902  & 0.365 & 0.813  & 0.367 & 0.815  & 0.530 & 0.841 \\
\langid{DE} & 0.581 & 0.943  & 0.355 & 0.839  & 0.313 & 0.929  & 0.388 & 0.949  & 0.266 & 0.868  & 0.454 & 0.959  & 0.392 & 0.913  & 0.393 & 0.914  & 0.564 & 0.926 \\
\langid{ES} & 0.524 & 0.927  & 0.296 & 0.815  & 0.260 & 0.902  & 0.323 & 0.875  & 0.229 & 0.811  & 0.333 & 0.929  & 0.331 & 0.876  & 0.327 & 0.876  & 0.490 & 0.888 \\
\langid{RU} & 0.535 & 0.871  & 0.470 & 0.770  & 0.327 & 0.845  & 0.417 & 0.887  & 0.313 & 0.791  & 0.472 & 0.864  & 0.428 & 0.838  & 0.422 & 0.838  & 0.597 & 0.850 \\
\langid{NL} & 0.543 & 0.939  & 0.292 & 0.815  & 0.274 & 0.887  & 0.366 & 0.912  & 0.265 & 0.865  & 0.409 & 0.943  & 0.360 & 0.892  & 0.358 & 0.893  & 0.536 & 0.905 \\
\langid{KO} & 0.445 & 0.916  & 0.401 & 0.812  & 0.321 & 0.938  & 0.403 & 0.935  & 0.220 & 0.835  & 0.362 & 0.945  & 0.359 & 0.896  & 0.359 & 0.897  & 0.529 & 0.900 \\
\langid{FA} & 0.498 & 0.870  & 0.386 & 0.759  & 0.450 & 0.884  & 0.327 & 0.788  & 0.202 & 0.641  & 0.399 & 0.822  & 0.361 & 0.790  & 0.377 & 0.794  & 0.535 & 0.816 \\
\langid{TR} & 0.459 & 0.900  & 0.338 & 0.823  & 0.295 & 0.898  & 0.403 & 0.892  & 0.274 & 0.849  & 0.376 & 0.944  & 0.361 & 0.884  & 0.358 & 0.884  & 0.538 & 0.893 \\
\langid{ZH} & 0.396 & 0.854  & 0.398 & 0.821  & 0.368 & 0.872  & 0.468 & 0.920  & 0.291 & 0.816  & 0.473 & 0.878  & 0.397 & 0.860  & 0.399 & 0.860  & 0.555 & 0.880 \\
\langid{HI} & 0.492 & 0.875  & 0.390 & 0.810  & 0.410 & 0.905  & 0.460 & 0.886  & 0.266 & 0.791  & 0.387 & 0.902  & 0.401 & 0.861  & 0.401 & 0.861  & 0.578 & 0.882 \\
\langid{BN} & 0.459 & 0.886  & 0.334 & 0.838  & 0.265 & 0.898  & 0.372 & 0.913  & 0.192 & 0.752  & 0.365 & 0.906  & 0.331 & 0.867  & 0.331 & 0.866  & 0.506 & 0.888 \\
\langid{MULTI} & 0.479 & 0.645  & 0.322 & 0.516  & 0.305 & 0.533  & 0.401 & 0.589  & 0.240 & 0.443  & 0.411 & 0.567  & 0.356 & 0.545  & 0.360 & 0.549  & 0.543 & 0.689 \\
\langid{MIX} & 0.517 & 0.792  & 0.308 & 0.685  & 0.324 & 0.687  & 0.387 & 0.722  & 0.235 & 0.563  & 0.421 & 0.739  & 0.364 & 0.695  & 0.365 & 0.698  & 0.577 & 0.828 \\

\bottomrule
    \end{tabular}}
    \caption{XLM-RoBERTa (B) baseline and {GEMNET} (G) domain results as measured by the F1 score for the different NER tags. The last three columns show the \emph{micro}, \emph{macro}, and \emph{mention detection} -- MD F1 performance. }
    \label{tab:domain_results}
\end{table*}

\end{document}